\documentclass[sigconf]{acmart}

\AtBeginDocument{%
  \providecommand\BibTeX{{%
    \normalfont B\kern-0.5em{\scshape i\kern-0.25em b}\kern-0.8em\TeX}}}

\setcopyright{acmlicensed}
\copyrightyear{2024}
\acmYear{2024}
\acmDOI{10.1145/3589335.3665995}

\acmConference[WWW '24 Companion] {Companion Proceedings of the ACM Web Conference 2024}{May 13--17, 2024}{Singapore, Singapore.}
\acmISBN{979-8-4007-0172-6/24/05}

\usepackage{enumitem}

\begin{document}

\title{OSPC: Detecting Harmful Memes with Large Language Model as a Catalyst}

\author{Jingtao Cao}
\authornote{Both authors contributed equally to this research.}
\email{jcao@se.cuhk.edu.hk}
\affiliation{%
  \institution{MoE Key Laboratory of High Confidence Software Technologies,\\
  The Chinese University of Hong Kong}
  \country{China}
}

\author{Zheng Zhang}
\authornotemark[1]
\email{cjang_cjengh@sjtu.edu.cn}
\affiliation{%
  \institution{The Hong Kong University of Science and Technology (Guangzhou)}
   \country{China}
}

\author{Hongru Wang}
\email{hrwang@se.cuhk.edu.hk}
\affiliation{%
  \institution{MoE Key Laboratory of High Confidence Software Technologies,\\
  The Chinese University of Hong Kong}
   \country{China}
}

\author{Bin Liang}
\email{bin.liang@cuhk.edu.hk}
\affiliation{%
  \institution{MoE Key Laboratory of High Confidence Software Technologies,\\
  The Chinese University of Hong Kong}
   \country{China}
}

\author{Hao Wang}
\email{haowang@hkust-gz.edu.cn}
\affiliation{%
  \institution{The Hong Kong University of Science and Technology (Guangzhou)}
   \country{China}
}

\author{Kam-Fai Wong}
\authornote{Corresponding Author.}
\email{kfwang@se.cuhk.edu.hk}
\affiliation{%
  \institution{MoE Key Laboratory of High Confidence Software Technologies,\\
  The Chinese University of Hong Kong}
   \country{China}
}







\begin{abstract}

Memes, which rapidly disseminate personal opinions and positions across the internet, also pose significant challenges in propagating social bias and prejudice. This study presents a novel approach to detecting harmful memes, particularly within the multicultural and multilingual context of Singapore. Our methodology integrates image captioning, Optical Character Recognition (OCR), and Large Language Model (LLM) analysis to comprehensively understand and classify harmful memes. Utilizing the BLIP model for image captioning, PP-OCR and TrOCR for text recognition across multiple languages, and the Qwen LLM for nuanced language understanding, our system is capable of identifying harmful content in memes created in English, Chinese, Malay, and Tamil. To enhance the system's performance, we fine-tuned our approach by leveraging additional data labeled using GPT-4V, aiming to distill the understanding capability of GPT-4V for harmful memes to our system. Our framework achieves top-1 at the public leaderboard of the Online Safety Prize Challenge hosted by AI Singapore, with the AUROC as 0.7749 and accuracy as 0.7087, significantly ahead of the other teams. Notably, our approach outperforms previous benchmarks, with FLAVA achieving an AUROC of 0.5695 and VisualBERT an AUROC of 0.5561.

\end{abstract}




\received{17 March 2024}
\received[revised]{30 May 2024}
\received[accepted]{23 May 2024}

\maketitle

\section{Introduction}

Memes, typically comprising an image with corresponding text, spread personal opinions and positions rapidly across the internet. However, this same characteristic makes them a powerful tool for disseminating social bias and prejudice. Such harmful memes can perpetuate stereotypes, foster discrimination, and exacerbate social divisions in a wide variety of social dimensions, including race, religion, sexual orientation, and more \cite{sharma2022detecting}. Recognizing this issue, AI Singapore has launched the Online Safety Prize Challenge \citep{lim2024ospc} to stimulate research of technologies that can effectively detect harmful memes, particularly in Singapore's multicultural and multilingual context. Given the country's linguistic diversity, these memes could be created and spread in various languages, including English, Chinese, Malay, and Tamil. This adds an additional layer of complexity to the detection process, as the memes not only embody the multimodal nature combining visual imagery with text to convey messages that are often nuanced and context-dependent but also do so across different cultural and linguistic contexts.

\begin{figure*}
  \includegraphics[width=0.9\linewidth]{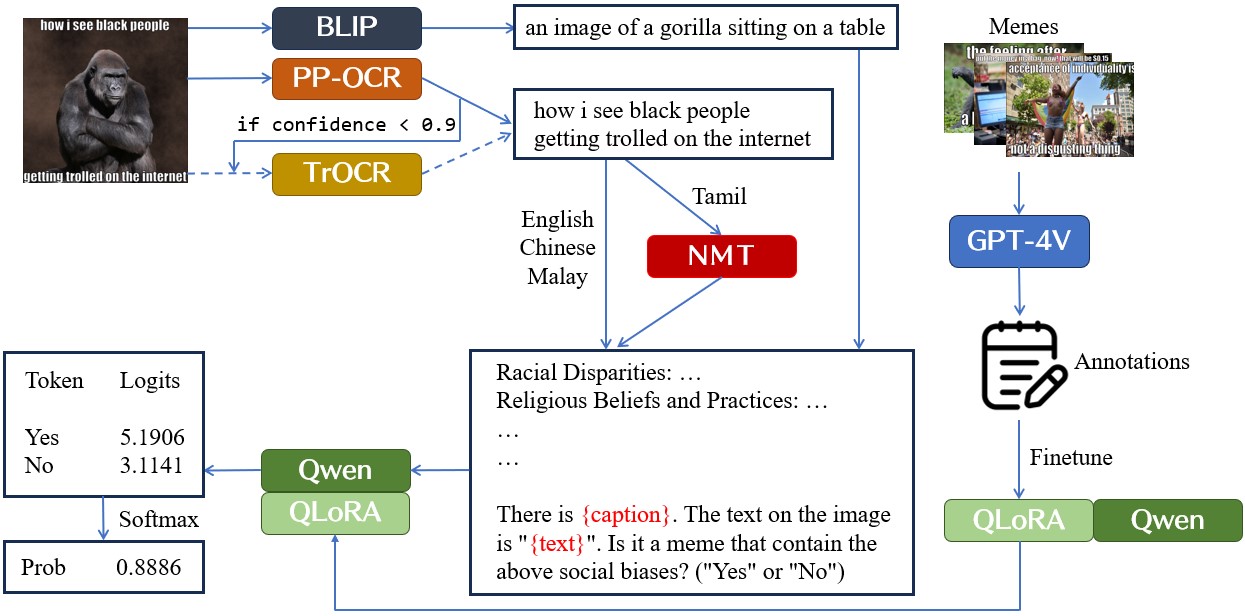}
  \caption{The system pipeline for meme analysis.}
  \Description{After LLM receiving a prompt that includes image caption and OCR text as input, we extract the logits for "Yes" and "No" for the next token and calculate the softmax for "Yes" to obtain the probability of the meme being harmful.}
  \label{fig:pipeline}
\end{figure*}

Most of previous methods either focus on single-modal information or overlook the nuanced context in a multilingual environment. First of all, traditional hate speech detection systems \cite{JAHAN2023126232, malik2023deep, 9696324}, primarily designed for textual analysis, fall short in interpreting the intricate interplay of visuals and text that memes embody. Secondly, most multimodal models trained for memes are frequently limited by their reliance on datasets with a predominantly Western context \cite{10.5555/3495724.3495944, hu2024visual, mei2023improving}, overlooking the cultural specificities and diverse linguistic nuances present in global online communities, and particularly in a linguistically rich and diverse country like Singapore.

To address these challenges, our solution leverages a composite system, integrating visual and textual analysis to understand and classify memes in a comprehensive way. As shown in Figure \ref{fig:pipeline}, the approach consists of three key steps: First, we employ a caption model to generate captions from meme images, extracting the underlying context and thematic elements of the visual content. Secondly, Optical Character Recognition (OCR) technology is utilized to identify and transcribe the text embedded within the memes, regardless of the language. Finally, we construct a prompt that incorporates both the caption and the text, which is then fed into a large language model (LLM), in order to get the probabilities of the next token (i.e, “Yes” or “No”) to assess if the meme is harmful due to social bias. In summary, our contributions can be summarized as follows:

\begin{itemize}[leftmargin=*]
    \item We propose a systematic framework to detect harmful multimodal memes, considering all features in different modals based on LLMs.

    \item To further improve the performance of each sub-module, we propose several tailored techniques, including automatic dataset collection method to build additional training corpus for better performance (e.g., OCR and LLM), and additional translation model to address poor performance at low-resource languages.

    \item Our framework achieve top-1 at the public leaderboard, with the AUROC as 0.7749 and accuracy as 0.7087, significantly ahead of the other teams.
\end{itemize}


\section{Methodology}

\subsection{Overall Framework}

In this section, we first introduce the overall framework of our method and then present details of each component. As shown in Figure \ref{fig:pipeline}, our method integrates a three-step process for meme analysis: 1) \textbf{Step one: multimodal features extraction}: we utilize existing off-the-shelf models to convert visual content to the textual descriptions, including a image-to-caption model (e.g., BLIP) and two OCR models (e.g., PP-OCR and TrOCR), similar with \citep{m3sum}; 2) \textbf{Step two: classification based on LLMs}: we then classify memes for harmful content using Qwen with a careful-designed prompts to incorporate all intermediate results in previous step. Furthermore, we utilize a translation model (e.g., Tamil to English) to better improve the performance of low-resource languages; 3) \textbf{Step three: fine-tuning with more training corpus}: we fintune our approach by leveraging additional data labeled using GPT-4V automatically, aiming to distill the understanding capability of GPT-4V for harmful memes to our system.

\subsection{Multimodal Feature Extraction}
\label{ocr}

To extract the visual context of memes, we utilize the BLIP model \cite{li2022blip} for image captioning, converting images into descriptive text to facilitate comprehensive meme analysis through LLMs. For OCR, the PP-OCR model \cite{li2022ppocrv3} is initially used due to its high speed, accuracy, and multilingual support, despite its limitations with Tamil text and complex training requirements. To overcome these, we also employ the TrOCR model \cite{10.1609/aaai.v37i11.26538}, a transformer-based OCR solution that offers simplicity in training data format, enhancing adaptability for diverse datasets.

\begin{figure}
  \includegraphics[width=\linewidth]{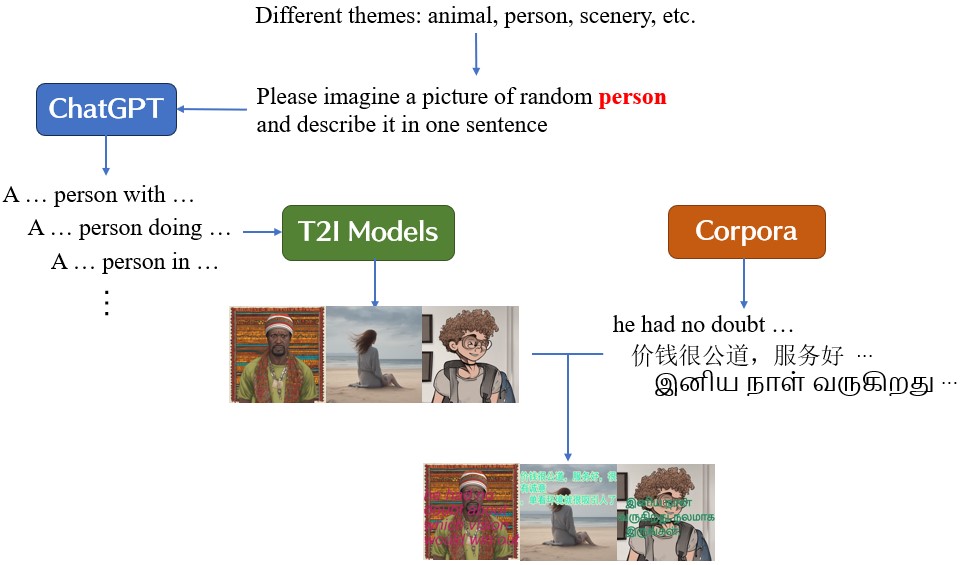}
  \caption{The construction method of OCR datasets.}
  \label{fig:ocrdata}
\end{figure}

As depicted in Figure \ref{fig:ocrdata}, to construct a robust and diverse OCR dataset, we employed ChatGPT \cite{openai2022chatgpt} to generate a wide range of Text-to-Image (T2I) prompts across various themes. 
These prompts were then used with Stable Diffusion \cite{rombach2022highresolution} and SDXL \cite{podell2023sdxl} to generate approximately 52,000 images. On these images, we printed text in random locations, using random fonts, sizes, and colors. The texts were sourced from corpora in English, Chinese, Malay, and Tamil, collected from the internet\footnote{Most of the corpus comes from \url{https://github.com/mesolitica/malaysian-dataset}, which contains a large amount of Malaysian internet corpus. Given 
the similar linguistic environment between Malaysia and Singapore, it includes content in English, Chinese, Malay, and Tamil.}. This approach allowed us to create a highly diverse OCR dataset, encompassing a wide range of languages and text appearances.

Our OCR strategy is designed to maximize accuracy and efficiency. Initially, we used PP-OCR to recognize text in memes. If the output confidence score is above 0.9, we accept the recognition result as it indicates high reliability. A confidence score below 0.9 usually suggests that the text is either in Tamil or is difficult to recognize for other reasons. In such cases, we switch to our trained TrOCR model for a second attempt at recognition. This dual-model approach ensures that we can accurately transcribe text from a wide variety of memes, thereby enhancing the overall performance of our meme analysis system.

\subsection{Classifying Memes with LLM}
\label{clallm}

For the final step, we employ Qwen \cite{qwen} as the base LLM, due to its commendable performance across multiple languages including English, Chinese, and Malay. Also, Its ability to understand various cultural contexts makes it an ideal candidate for our purposes. Given the constraints on computational speed and deployment size, we opt for the Qwen1.5-14B-Chat-Int4 variant. Recognizing Qwen's limitations in handling Tamil, we supplement it with a Tamil to English Transformer translation model \cite{NIPS2017_3f5ee243}, specifically trained for this task. The training corpus for this model was derived from translations of the Tamil corpus mentioned in section \ref{ocr}, facilitated by ChatGPT.       

Upon obtaining the caption and text of a meme, if the text is in Tamil, it is first translated into English using our translation model. The translated or original text, along with the caption, is then fed into the following prompt template:

\begin{quote}
Racial Disparities: Memes perpetuating stereotypes or prejudices based on race or ethnicity.\\
Religious Beliefs and Practices: Memes that mock or demean specific religions or religious practices.\\
Sexual Orientation: Memes that promote negative stereotypes or biases about different sexual orientations.\\
Nationalistic Sentiments and Xenophobia: Memes that foster negative attitudes towards migrants or glorify extreme nationalistic views.\\
Socio-Economic Divides: Memes that highlight or ridicule class struggles and socio-economic disparities.\\
Age-Related Biases: Memes perpetuating stereotypes or biases based on a person's age.\\
Gender Discrimination: Memes that promote gender stereotypes or discriminate based on gender.\\
Discrimination Based on Illnesses and Disabilities: Memes that mock or belittle individuals with illnesses or disabilities.\\

There is \{caption\}. The text on the image is "\{text\}". Is it a meme that contain the above social biases? ("Yes" or "No")
\end{quote}

This prompt template clearly delineates various definitions of harmful memes, thereby aiding the LLM in better identifying the harmful nature of a meme. After the LLM processes this prompt, we extract the logits corresponding to the “Yes” and “No” tokens from the next token logits computed by the LLM. We then calculate the probability of “Yes” using the following formula, which represents the final probability of the meme being harmful:

\begin{equation}
P(\text{Harmful}) = P(\text{"Yes"}) = \frac{e^{\text{logit}(\text{"Yes"}) / t}}{e^{\text{logit}(\text{"Yes"}) / t} + e^{\text{logit}(\text{"No"}) / t}}
\end{equation}

where $t$ is a temperature parameter. This formula quantifies the likelihood of a meme being harmful based on the LLM's analysis, enabling a systematic and nuanced approach to detecting harmful content within memes.

\subsection{Fine-Tuning Process}

To enhance Qwen's understanding of the harmfulness of memes, we fine-tuned the Qwen1.5-14B-Chat-Int4 using QLoRA \cite{dettmers2023qlora}. We utilized GPT-4V \cite{openai2023gpt4}, the vision version of GPT-4 that can accept images as input and has a strong understanding and grasp of various linguistic and cultural backgrounds, to annotate the memes in the Hateful Memes Challenge Dataset \cite{10.5555/3495724.3495944}. For each meme, GPT-4V determined whether it was harmful and provided corresponding reasons for this judgment.

The dataset is in a chat style. The questions were structured in a manner similar to the prompt in section \ref{clallm}. The responses consisted of a simple “Yes” or “No” followed by a narrative explaining the reason behind the judgment.

\section{Discussions}

In addressing the challenge of detecting harmful memes within the multicultural and multilingual context of Singapore, our solution primarily optimized for English memes. Given the multilingual and multicultural knowledge base of Qwen, we hypothesized that improvements made in detecting harmful English memes could be transferable to memes in other languages and cultural contexts. However, this assumption proved difficult to empirically validate due to limitations in data constraints.

Besides, in the process of cleaning and annotating data, we employed GPT-4V, one of the most powerful multilingual and multimodal LLMs available. While GPT-4V demonstrated strong performance in English, Chinese, and Malay, its capabilities in Tamil were notably weaker, likely due to the scarcity of Tamil language content on the internet. This observation underscores the ongoing challenges in leveraging LLMs for low-resource languages like Tamil.

\section{Limitations}

Our approach to detecting harmful memes relies on converting the visual information present in memes into text, which is then analyzed by a LLM. However, we observed that the descriptions of some memes generated by BLIP were overly simplistic or failed to recognize specific historical figures in memes. This limitation led to inaccuracies in the subsequent analysis of memes by the LLM.

Moreover, to address the deficiency of Qwen in processing Tamil, we incorporated a translation model to convert Tamil text into English before analysis. Ideally, enhancing Qwen's capabilities to directly understand and analyze Tamil without the need for translation would be preferable. This 
would not only streamline the process but also potentially improve the accuracy of meme analysis by preserving the original nuances of the language. However, achieving this level of linguistic proficiency in Qwen, especially for a low-resource language like Tamil, remains a challenge we have yet to overcome.

\section{Conclusion and Future Work}

Our approach to detecting harmful memes through a combination of image captioning, OCR, and LLM analysis represents a significant step forward in addressing the challenges posed by the multimodal and multilingual nature of memes. By leveraging models like BLIP for image captioning, PP-OCR and TrOCR for text recognition, and Qwen for nuanced language understanding, we have developed a system capable of identifying harmful content across a variety of languages and cultural contexts. To enhance the system's ability, we fine-tuned Qwen and aligned its understanding of meme harmfulness with GPT-4V.

For future work, we aim to explore the potential of multimodal LLMs that could offer a more integrated approach to analyzing the visual and textual components of memes. Additionally, enhancing the system's capability to directly process and understand all languages present in the memes, including low-resource ones like Tamil, without relying on translation, will be a key focus. By addressing these challenges, we hope to further refine our solution, making it more effective and efficient in creating safer online environments by mitigating the spread of harmful content.



\bibliographystyle{ACM-Reference-Format}
\bibliography{sample-base}

\appendix

\end{document}